\documentclass[11pt]{article}
\usepackage{graphicx}
\usepackage{amssymb}
\usepackage{amscd}
\usepackage{mathrsfs}
\usepackage{longtable,lscape}
\usepackage{amsthm}
\usepackage{amsfonts}
\usepackage{amsmath}
\usepackage{bbm}
\usepackage{float}
\usepackage{url}

\newcommand{\captionfonts}{\footnotesize}
\makeatletter  
\long\def\@makecaption#1#2{%
  \vskip\abovecaptionskip
  \sbox\@tempboxa{{\captionfonts #1: #2}}%
  \ifdim \wd\@tempboxa >\hsize
    {\captionfonts #1: #2\par}
  \else
    \hbox to\hsize{\hfil\box\@tempboxa\hfil}%
  \fi
  \vskip\belowcaptionskip}
\makeatother 
\begin{document}
\title{Testing Quantum Models of Conjunction Fallacy \\ on the World Wide Web}
\author{Diederik Aerts$^1$, Jonito Aerts Argu\"elles$^2$, Lester Beltran$^3$, Lyneth Beltran$^1$, \\ Massimiliano Sassoli de Bianchi$^{1}$, Sandro Sozzo$^{4}$  and Tomas Veloz$^1$ \vspace{0.5 cm} \\ 
        \normalsize\itshape
        $^1$ Center Leo Apostel for Interdisciplinary Studies, 
         Brussels Free University \\ 
        \normalsize\itshape
         Krijgskundestraat 33, 1160 Brussels, Belgium \\
        \normalsize
        E-Mails: \url{diraerts@vub.ac.be},\\\url{lyneth.benedictinelawcenter@gmail.com},\\\url{tveloz@gmail.com}
          \vspace{0.5 cm} \\ 
        \normalsize\itshape
        $^2$ KASK and Conservatory, \\
        \normalsize\itshape
         Jozef Kluyskensstraat 2, 9000 Ghent, Belgium
        \\
        \normalsize
        E-Mail: \url{jonitoarguelles@gmail.com}
	  \vspace{0.5 cm} \\ 
        \normalsize\itshape
        $^3$ 825-C Tayuman Street, Tondo \\
        \normalsize\itshape
        Manila, The Philippines
        \\
        \normalsize
        E-Mail: \url{lestercc21@gmail.com}
          \vspace{0.5 cm} \\ 
        \normalsize\itshape
        $^4$ School of Management and IQSCS, University of Leicester \\ 
        \normalsize\itshape
         University Road, LE1 7RH Leicester, United Kingdom \\
        \normalsize
        E-Mail: \url{ss831@le.ac.uk} 
       	\\
              }
\date{}
\maketitle
\begin{abstract}
\noindent
The `conjunction fallacy' has been extensively debated by scholars in cognitive science and, in recent times, the discussion has been enriched by the proposal of modeling the fallacy using the quantum formalism. Two major quantum approaches have been put forward: the first assumes that respondents use a two-step sequential reasoning and that the fallacy results from the presence of `question order effects'; the second assumes that respondents evaluate the cognitive situation as a whole and that the fallacy results from the `emergence of new meanings', as an `effect of overextension' in the conceptual conjunction. Thus, the question arises as to determine whether and to what extent conjunction fallacies would result from `order effects' or, instead, from `emergence effects'. To help clarify this situation, we propose to use the World Wide Web as an `information space' that can be interrogated both in a sequential and non-sequential way, to test these two quantum approaches. We find that `emergence effects', and not `order effects', should be considered the main cognitive mechanism producing the observed conjunction fallacies.
\end{abstract}

\section{Introduction}\label{intro}

Since the early pioneering studies, there has been growing interest in the field of investigation originally known as `quantum cognition', with the aim of better and more thoroughly understanding the extended logic and mechanisms governing our human cognitive processes \cite{aa1995,a2009a,ys2010,bb2012,abgs2013,ags2013,hk2013,pb2013}. Among the situations that have been more extensively studied, there is the so-called `conjunction fallacy', firstly identified by Tversky and Kahneman in their now famous `Linda story' \cite{tk1983}, which stimulated an intense debate within the cognitive science community \cite{tk1983,mb1984,g1996,tbo2004,mo2009}. 

In the conjunction fallacy, people judge the conjunction of two events as more likely than one of the two events taken separately. The fallacy has now been clearly observed in many experiments involving different Linda-like stories, and situations of `double fallacies' have also been identified, where the conjunction is considered to be more likely than both individual events forming the conjunction \cite{mb1984,tbo2004,gre1991,fp1996,f2002,wm2008,c2009,Lu2015}.

Typically, two `quantum cognition approaches' have been used so far to explain and model the conjunction fallacy: the first is based on `sequential measurements' and `question order effects' \cite{ys2010,bb2012,hk2013,bpft2011}, whereas the second is based on the `modeling of concepts as quantum entities in different states', and the `emergence of new meanings when concepts are combined' \cite{ys2010,bb2012,hk2013,bpft2011,abssv2016}.\footnote{Emergence of new meanings in conceptual combinations closely resembles the `holistic quantum computational semantics' developed by Dalla Chiara {\it et al.}, where the meaning of a sentence is not attributed to the compositional meaning of each word but, rather, to the holistic relations between these words (see, e.g., \cite{Dallachiaraetal2006,Dallachiaraetal2016}).} For the sake of simplicity, we denote the first approach the `order effects' (OE) modeling, and the second approach the `emergence effects' (EE) modeling. 

These two approaches are not necessarily incompatible, as is clear that they describe different ways participants may respond to a Linda-like interrogative context. More precisely, given a certain number of participants, for one portion of them the cognitive situation might be predominantly evaluated by means of a sequential reasoning, and therefore OE are most important to model their action, whereas for another portion a direct evaluation of the conceptual combination expressed by the conjunction might be predominant, and therefore EE are most important to model their action.
Of course, more complex situations can also be imagined, but for the sake of simplicity we assume that the cognitive processes described by the OE and EE models are the main ones. 

Without additional information, it is difficult to ascertain, on the basis of the sole statistics of outcomes, if either sequential or emergent reasoning are mostly applied by the respondents, or if both are used by a comparable number of them. However, there are two important clues already suggesting that OE are not sufficient, alone, to generally explain the observed conjunction fallacies. The first one is that OE, contrary to EE, do not allow for the modeling of double conjunction fallacies. The second one is that there are experimental situations exhibiting a conjunction fallacy but for which one can show, through additional investigation, that the OE they produce are insufficient to explain the fallacy as resulting from a sequential reasoning \cite{bketal2015}.

The main purpose of the present paper is to use that specific `information space' known as the World Wide Web (WWW) to provide evidence that EE, as described in Brussels' operational-realistic (quantum-based) approach to cognition \cite{a2009a,abssv2016,asdbs2016}, are most probably the dominant ingredient in the observed conjunction fallacies, as possible contributions from OE are generally of a lower order of magnitude as compared to those from EE. To this end we proceed as follows. 

In Sect.~\ref{ordereffects}, we briefly recall the quantum modeling in terms of sequential measurements, emphasizing that it cannot be used to model double conjunction fallacies. In Sect.~\ref{emergenteffect}, we also briefly recall the quantum modeling of the conjunction fallacy in terms of superposition and interference effects resulting from concepts' combinations, when the latter are described as `meaning entities that can be in different states'. 

Following the approach presented in \cite{acds2010,a2011}, we then show in Sect.~\ref{WWW} how to interrogate the WWW to highlight the presence of conjunction fallacies. In addition, we show that, similarly to the quantum modeling, sequential interrogations cannot produce double fallacies, but also that EE generally produce overextensions of probabilities that are typically one or more orders of magnitude larger than those possibly associated with OE.

In our analysis, we make use of the key notion of `meaning bond', as defined in \cite{a2011}, and two paradigmatic examples are also considered in Sec.~\ref{Guppy}, as a means to illustrate our results: the `Pet-Fish problem' and the `Linda story'. Finally, in Sect.~\ref{Conclusion}, we offer a few concluding remarks.

\section{The conjunction fallacy as a question order effect}\label{ordereffects}
We briefly explain here why the modeling of the conjunction fallacy in terms of interference resulting from sequential non-compatible measurements  \cite{bb2012,bpft2011,pnas} cannot account for situations where a double conjunction fallacy is observed. 

In this approach, one first assumes that the situation under study generates a `state of belief', represented by a vector $|S\rangle$ of a Hilbert space ${\cal H}$. Then, one interprets the conjunction `$A$ and $B$' as the sequence `$A$ then $B$' (with $A$ for instance the more likely of the two events), and to model the associated probability one introduces two orthogonal projection operators $M_A$ and $M_B$, representing the `yes-no questions' that participants are assumed to answer, sequentially (in their minds), when judging the conjunction `$A$ and $B$'. 

The probability $p(A)$ of answering `yes' the first question is then $p(A)= \| M_A|S\rangle\|^2$, which produces the revised state $|S_A\rangle ={M_A|S\rangle \over \| M_A|S\rangle\|}$. Similarly, the probability $p(A')$ of answering `no' the first question is $p(A')= \| M'_A|S\rangle\|^2$, which produces the revised state $|S_{A'}\rangle ={M'_A|S\rangle \over \| M'_A|S\rangle\|}$, where we have defined the projection $M'_A=\mathbbmss{1}-M_A$. On the other hand, the probability of obtaining the answer `yes' for the second $B$-question, irrespective of the answer given to the first $A$-question, is $p(A\ {\rm then}\ B) + p(A'\ {\rm then}\ B)=p(A) p(B|A) + p(A')p(B|A')$, 
where $p(B|A)$ (resp., $p(B|A')$) is the conditional probability of answering `yes' the second question, knowing the answer to the first question was `yes' (resp., `no'). 
So, we have: $p(A\ {\rm then}\ B) + p(A'\ {\rm then}\ B )= \| M_A|S\rangle\|^2\, \| M_B|S_{A}\rangle\|^2 + \| M'_A|S\rangle\|^2\, \| M_B|S_{A'}\rangle\|^2= \| M_BM_A|S\rangle\|^2+\| M_BM'_A|S\rangle\|^2$. Also, the probability $p(B)$ of answering `yes' to the $B$-question, when asked first, is $p(B)= \| M_B|S\rangle\|^2$, and after some simple algebra one obtains: 
\begin{equation}
p(A\, {\rm then}\, B) - p(B) = -[p(A'\, {\rm then}\, B)+ {\rm Int}_B],
\label{conditionx}
\end{equation}
where the so-called interference contribution  ${\rm Int}_B$ is given by: 
\begin{eqnarray}
{\rm Int}_B&=&\| M_B|S\rangle\|^2 - \| M_BM_A|S\rangle\|^2-\| M_BM'_A|S\rangle\|^2\nonumber\\
&=& \langle S | (M_A+M'_A)M_B(M_A+M'_A) |S\rangle \nonumber\\
&&\quad - \langle S |M_AM_BM_A|S\rangle - \langle S |M'_AM_BM'_A|S\rangle \nonumber\\
&=&\langle S |M_AM_BM'_A|S\rangle + \langle S |M'_AM_BM_A|S\rangle.
\label{conditionx2}
\end{eqnarray}
Note that ${\rm Int}_B$ is non-zero only if $[M_A,M_B]\neq 0$, i.e., if the two questions are incompatible and therefore give rise to OE. 

If `$A$ and $B$' is interpreted as the sequence `$A$ then $B$', it is clear that we can model a conjunction fallacy if the term in brackets in (\ref{conditionx}) is strictly negative, i.e., if ${\rm Int}_B<-p(A'\, {\rm then}\, B)$, which can certainly be the case for suitable choices of $|S\rangle$ and of the projections $M_A$ and $M_B$. On the other hand, the reason why we cannot model a double conjunction fallacy in this way, i.e., that we cannot also satisfy the inequality $p(A\, {\rm then}\, B) < p(A)$, is that since $A$ is assumed to be answered first, we have $p(A) = p(A\, {\rm then}\, B) + p(A\, {\rm then}\, B')$, as is clear that: $p(A)= \| M_A|S\rangle\|^2=\langle S |M_A(M_B+M'_B)M_A|S\rangle=\| M_BM_A|S\rangle\|^2+\| M'_BM_A|S\rangle\|^2$. Thus, we have $p(A\, {\rm then}\, B) - p(A) = -p(A\, {\rm then}\, B')\leq0$,
which means that we can never describe a double conjunction fallacy (a double overextension of the probabilities) by means of sequential measurements.

However, as emphasized in Sect. \ref{intro}, experimental situations have been identified which exhibit a double conjunction fallacy. Thus, considering that a quantum modeling in terms of sequential measurements is structurally unable to model these situations, this leads some doubts that conjunction fallacies should be generally accounted for in terms of OE. Also, some authors have recently identified experimental situations where conjunction fallacies are present but order effects are statistically insignificant \cite{bketal2015}.

\section{The conjunction fallacy as an emergence effect}\label{emergenteffect}

We briefly explain here how, based on Brussels' operational-realistic approach to cognition \cite{asdbs2016}, a conjunction fallacy (be it single or double) can be modeled as resulting from `emergent reasoning' in the ambit of a single (instead of sequential) measurement context, with the fallacy being described as a `superposition effect' due to the combination of two (or more) concepts. 

In this approach human concepts are described as `entities in well-defined states', rather than `containers of instantiations', as is done in traditional set-theory based concept theories. This means that during the interaction with a context, like for instance an interrogative (measurement) context, the state of a conceptual entity can change. For the sake of brevity, we do not dwell here on the details of Brussels' approach to cognition and refer the interested reader to \cite{asdbs2016,IQSA2} and references therein. We just present in the following the gist of our approach and the ensuing modeling.

Unlike the approach of Sect. \ref{ordereffects}, the Hilbert space structure is here used not to represent the `states of belief' but, rather, the `states of the conceptual entities' involved in the cognitive situation under study. Thus, $A$ and $B$ now denote concepts (i.e., human conceptual entities) whose states we represent by the unit vectors $|A\rangle$ and $|B\rangle$, respectively, of a Hilbert space ${\cal H}$. Then, the state of a concept that is a combination of $A$ and $B$, like the conjunction `$A$ and $B$', is represented by the linear combination $|A \ \textrm{and} \ B\rangle=\frac{1}{\sqrt{2}}(|A\rangle+|B\rangle)$ in ${\cal H}$. In other terms, the model assumes that the new concept `$A$ and $B$', resulting from the conjunction of $A$ and $B$, emerges from the latter exactly as the linear superposition of two quantum states emerges from its component states (here assumed to be orthogonal).

An orthogonal projection operator $M$ is then  assumed to represent the `yes-no question' that the participants have to answer, in relation to concepts $A$, $B$ and of their possible combinations, here the conjunction `$A$ and $B$'. For instance, the question can be that of knowing if another given concept $C$ is considered to be a typical representative of $A$, $B$ and `$A$ and $B$'. The probability of obtaining the answer `yes' to these three questions is then given by the averages $p(A)=\langle A|M|A\rangle$, $p(B)=\langle B|M|B\rangle$ and $p(A\, {\rm and}\, B) = \langle A \, \textrm{and} \, B|M |A \, \textrm{and} \, B \rangle=\frac{1}{2}[p(A)+p(B)]+\Re\langle A|M|B\rangle$, where the real part of $\langle A|M|B\rangle$, denoted $\Re\langle A|M|B\rangle$, is the so-called `interference term'. Writing
\begin{eqnarray} 
&p(A \, \textrm{and} \, B)-p(A) =\frac{1}{2}\left[p(B)-p(A)\right]+\Re\langle A|M|B\rangle,\label{AND2a}\\
&p(A \, \textrm{and} \, B)-p(B) =-\frac{1}{2}\left[p(B)-p(A)\right]+\Re\langle A|M|B\rangle,
\label{AND2b}
\end{eqnarray} 
we see from (\ref{AND2a}) that there can be a conjunction fallacy with respect to $A$ if $\Re\langle A|M|B\rangle \geq -\frac{1}{2}[p(B)-p(A)]$. 
Analogously, we see from (\ref{AND2b}) that there can be a conjunction fallacy with respect to $B$ if $\Re\langle A|M|B\rangle \geq \frac{1}{2}[p(B)-p(A)]$. Since these two conditions can be jointly satisfied if $\Re\langle A|M|B\rangle \geq \frac{1}{2}|p(B)-p(A)|$, the approach allows also for the modeling of situations of double conjunction fallacies (double overextensions of the joint probability), consistently with what is observed in certain experiments. 

It is worth mentioning that the approach we have here just outlined is in fact much more general, as it allows for the modeling of not only conjunctions effects, but also other cognitive effects in membership judgments, like deviations from classicality in disjunctions and negations \cite{a2009a,PhilTransA2015,asv2015,Sozzo2014}.

We also stress that, consistently with the prescriptions of Brussels' operational-realistic approach \cite{asdbs2016}, conjunction fallacies are now modeled not as OE, as was done in Sect. \ref{ordereffects}, but as EE, interpreting them as `effects of conceptual overextension resulting from interference effects associated with single (non-sequential) measurement situations'.

\section{The WWW as a ``story teller''}\label{WWW}
We provide in this section additional arguments in favor of the thesis that conjunction fallacies result mainly as the consequence of the emergence of new meaning, rather than the effect of incompatible questions. We do this by considering a situation which is related, although not completely equivalent, to that considered in concept research with respect to typicality and membership. The situation in question consists in interrogating that specific `information space' known as the World Wide Web (WWW), whose semantic structure results from the combined cognitive activity of a great number of human beings. 

More precisely, following \cite{acds2010,a2011}, we consider the WWW as a `mind-like entity' that can be interrogated by evaluating the number of webpages containing certain terms. Thus, instead of asking human participants to evaluate `meaning relations' (like typicality, membership, representativeness, likeliness, etc.) of certain concepts with respect to other concepts and their combinations, we directly evaluate them by counting the number of pages containing the terms associated with these concepts and combinations of concepts. 

Of course, using the WWW means working primarily with words and combinations of words, whereas human minds certainly deal with schemes of concepts whose structure is more complex than that of the WWW and its pages. The latter is however a reflection of our human conceptual activity, and it is reasonable to assume that the cognitive effects that can be identified in it are significantly related to those identified in cognitive research. When a concept is evaluated in relation to another concept, it is indeed very plausible that this evaluation will depend on how both concepts are connected in meaning. For this reason, we will introduce in the next section the notion of `meaning bond' between two terms of the WWW, which will give us a tool to express how concepts corresponding to these terms are connected in meaning.

\subsection{Meaning bond}

Let $n_A$ be the number of webpages containing the term $A$, and $n_W$ the total number of available webpages. The ratio 
\begin{equation}
p(A)={n_A\over n_W}
\label{ratio1}
\end{equation}
is then the probability of ``landing'' on a webpage containing the term $A$, when a webpage is randomly selected. 

If the WWW is viewed as a mind-like entity, then $p(A)$ can also be understood as the probability that the term $A$ would be present in a story that such mind would tell, when asked to tell an arbitrary (i.e., not predetermined) story. Here we are considering that each webpage is a possible story that the WWW can tell, and according to our operational-realistic approach to cognition \cite{asdbs2016} each story containing $A$ is a possible `state of $A$', i.e., the state defined by the specific semantic context of that story (i.e., webpage). 

In a similar way, we can define the probabilities $p(B)={n_B\over n_W}$ and $p(A,B)={n_{A,B}\over n_W}$, for the WWW to tell a story that is a `state of $B$', and jointly a `state of $A$' and a `state of $B$'. The `meaning bond' between $A$ and $B$ can then be defined as the ratio \cite{a2011}
\begin{equation}
M(A,B)={p(A,B)\over p(A)p(B)}={n_{A,B}\, n_W\over n_A\, n_B},
\label{bond1}
\end{equation}
with $0\leq M(A,B)<\infty$. If $A$ and $B$ are positively correlated, we have $M(A,B)>1$ and the meaning bond is said to be `attractive'. If $A$ and $B$ are negatively correlated, $M(A,B)<1$ and the meaning bond is said to be `repulsive', and when there is absence of correlation, we have $M(A,B)=1$, a situation that can be interpreted as `meaning neutrality'. 

To better understand the content of (\ref{bond1}), let us introduce the (conditional) probability $p(A|B)$ of ``landing'' on a webpage containing the term $A$, when a webpage is randomly selected among those containing the term $B$. This can also be understood as the probability that the WWW would arbitrarily tell a story which is a `state of $A$', when asked to tell a story that is a `state of $B$'. This probability is given by the ratio
\begin{equation}
p(A|B)={n_{A,B}\over n_B}.
\label{ratio2}
\end{equation}
Therefore, (\ref{bond1}) can also be written as
\begin{equation}
M(A,B)={p(A|B)\over p(A)}={P(B|A)\over p(B)}.
\label{bond1-bis}
\end{equation}
Thus, $M(A,B)$ is a ratio that expresses the relative frequency of appearance of $A$ (resp. $B$) in stories that are `states of $B$' (resp., `states of $A$'), as compared to its overall frequency of appearance in all stories (webpages) of the WWW. 

It is worth mentioning that in the definition of $M(A,B)$, webpages are counted without entering into the merits of the different possible semantic relations between $A$ and $B$, in the different semantic contexts of the webpages where they appear. This means that $M(A,B)$ is to be interpreted as a `first-order' approximation and foundation for a meaning bond measure. Indeed, in the definition of  $M(A,B)$, each webpage is given the same weight, while a more detailed inspection of the meaning relevance of webpages, could lead to attributing weights that take into account such individual-webpage-linked meaning relevance. We however have no straightforward tool available to get access to such more specific meaning relevance of individual webpages, so that the simple co-occurrence of terms with equal weights for each webpage is taken as the foundation for our meaning bond measure.

\subsection{Overextension effects}

In addition to the two terms $A$, $B$, we now also consider their conjunction $AB$, and a third term $C$. We want to compare $p(C|AB)$ with $p(C|A)$ and $p(C|B)$, interpreting these probabilities as an estimation of how human subjects evaluate (in a scale ranging from $0$ to $1$) a meaning relation (like typicality, membership, representativeness, likeliness, etc.) between $C$ and $AB$, $A$ and $B$, respectively. This means that the situations $p(C|AB)>p(C|A)$ and/or $p(C|AB)>p(C|B)$ are considered to be indicative of overextension effects (and more specifically of conjunction fallacies), due to the emergence of new meanings resulting from the  $AB$-combination. Considering for instance the first of these two inequalities, according to (\ref{bond1-bis}) we can also write
\begin{equation}
{p(C|AB)\over p(C|A)}= {M(C,AB)\over M(C,A)}={n_{AB,C} \over n_{A,C}}{n_{A} \over n_{AB}}>1,
\label{fallacy1}
\end{equation}
and similarly for the second inequality.

Thus, the notion of meaning bond can be used to explain (and predict) the existence of a conjunction fallacy, in the sense that if a term $C$ has a stronger meaning bond with respect to the conjunction $AB$ of two terms $A$ and $B$, than with respect to one of them, or both, taken individually, an effect of (single or double) overextension will necessarily appear.

Let us now also consider the possibility of obtaining an overextension effect as a result of OE. To this end we consider a sequence of two processes. The first consists in asking the WWW to tell a story that is a `state of $A$' and then checking if it is also a `state of $C$'. The second one is about asking the WWW to tell a story that is both a `state of $A$' and a `state of $C$', and then checking if it is also a `state of $B$'. Clearly, the probability $p(C|A\, {\rm then}\, B)$ of obtaining a `state of $C$' in the first process and then a `state of $B$' in the second process is given by the product
\begin{equation}
p(C|A\, {\rm then}\, B)= p(C|A) p(B|A,C) = {n_{A,C}\over n_A}\,{n_{A,B,C}\over n_{A,C}}={n_{A,B,C}\over n_A},
\label{conditional2}
\end{equation}
where $n_{A,B,C}$ is the number of webpages containing the three terms $A$, $B$ and $C$. Inverting the roles of $A$ and $B$, one obtains in a similar way that 
$p(C|B\, {\rm then}\, A)={n_{A,B,C}\over n_B}$.

The sequence `$A$ then $B$' can thus produce an overextension if we have that $p(C|A\, {\rm then}\, B)>p(C|B)$, or $p(C|A\, {\rm then}\, B)>p(C|A)$. For the first inequality, we can equivalently write: 
\begin{equation}
{p(C|A\, {\rm then}\, B)\over p(C|B)}= {p(C|B\, {\rm then}\, A)\over p(C|B)} \,{p(C|A\, {\rm then}\, B)\over p(C|B\, {\rm then}\, A)} >1.
\label{ordereffectoverextension}
\end{equation}
The first factor on r.h.s. of (\ref{ordereffectoverextension}) is ${n_{A,B,C}\over n_{B,C}}$, and since $n_{A,B,C}\leq n_{B,C}$, it is always less than $1$ (no overextension). Thus, the necessary condition for (\ref{ordereffectoverextension}) to express an overextension is that the second OE factor, which is equal to ${n_B\over n_A}$, is strictly greater than $1$. Similarly, for $p(C|B\, {\rm then}\, A)$, there can be an overextension only if ${n_A\over n_B}$ is strictly greater than $1$.

Thus, in accordance with the OE quantum model, and for the same structural reasons, we find that sequential processes can only generate single overextension effects.

\subsection{Comparing the EE and OE models}\label{comparing}

In the previous section, we have considered two different mechanisms for the generation of overextension effects (conjunction fallacies), analyzed from the WWW perspective: according to the first, the effects that are due to the emergent meanings of the combined concept $AB$, when considered as an indivisible whole, as expressed by (\ref{fallacy1}), and according to the second one the effects result from the presence of order effects, when a sequential process is considered, as expressed by (\ref{ordereffectoverextension}). 

We want now to compare these two mechanisms, to show that the EE will generally dominate the OE in the production of possible overextensions effects. This will be the case if
\begin{equation}
{p(C|AB)\over p(C|A\, {\rm then}\, B)}={n_{AB,C}\over n_{A,B,C}}{n_A\over n_{AB}}\gg 1,
\label{condition-1}
\end{equation} 
where the symbol ``$\gg$" means here that the above ratio is approximately of the order of magnitude of $10$. To investigate if this condition is generally fulfilled in relevant cognitive situations, we write $n_A=n_{A,\bar{B}}+n_{A,B}$, where $n_{A,\bar{B}}$ is the number of webpages containing $A$ but not $B$. The above condition then becomes
\begin{equation}
\left({n_{AB,C}\over n_{AB}}{n_{A,B}\over n_{A,B,C}}\right)\left(1+{n_{A,\bar{B}}\over n_{A,B}}\right)\gg 1.
\label{condition}
\end{equation}

The first factor in (\ref{condition}) is nothing but the ratio ${M(C,AB)\over M(C; A,B)}$ of the meaning bond between $C$ and the conjunction $AB$, and the meaning bond between $C$ and the two terms $A$ and $B$, when not necessarily manifesting as a strict $AB$-combination. Since, by hypothesis, we are in a situation where there are some overextension effects (whose strengths we want precisely to compare), we will generally have either ${M(C, AB)\over M(C; A,B)}\approx 1$, if the meaning associated with the joint presence of $A$ and $B$ can emerge even though the two terms do not manifest in a strict $AB$-combination, or ${M(C, AB)\over M(C; A,B)}\gg 1$, if such meaning (responsible for an increased bond with $C$) can instead only manifest when the two terms manifest one immediately after the other.  

Another way to say the same thing is to observe that since $n_{AB,C}=n_{AB,A,B,C}$ and $n_{AB}=n_{AB,A,B}$ (as is clear that a webpage containing the combination $AB$ also contains both $A$ and $B$), the first factor in (\ref{condition}) can also be written as the ratio ${p(AB|A,B,C)\over p(AB|A,B)}$, i.e., as the ratio between the relative frequency of appearance of the conjunction $AB$ in webpages containing the terms $A$, $B$ and $C$, and the relative frequency of appearance of the conjunction $AB$ in pages containing the terms $A$ and $B$. If $C$ does not entertain an additional meaning relation with $A$ and $B$, when they are in the strict combination $AB$, then we have ${p(AB|A,B,C)\over p(AB|A,B)}\approx 1$, and if it does, we have: ${p(AB|A,B,C)\over p(AB|A,B)}\gg 1$.

Thus, the first factor in (\ref{condition}) can be expected to be approximately equal to $1$, or much greater than $1$, depending on whether it is just the joint presence of $A$ and $B$ the cause for the EE, or the latter are due to their presence in the more strict $AB$ configuration. 

Let us now consider the second factor in (\ref{condition}). For this, we observe that since the WWW is a large collection of all sort of stories (webpages), if neither $A$ nor $B$ are so-called `stop words', which are almost present in all webpages (these are the most common words in a language, usually filter out in natural language processing and having a neutral meaning bond with all other terms), then we will generally have $n_{A,\bar{B}}\gg n_{A,B}$. To explain why this can be expected to be the case, let us present an heuristic reasoning.

Since we have assumed that $A$ and $B$ are not `stop words', there will be terms $D_i$, $i=1,\dots n$, almost not sharing any webpage with $B$, i.e., such that $n_{B,D_i}\approx 0$, for all $i$. We thus have: $n_{A,\bar{B}}\gtrsim \sum_{i=1}^n n_{A,D_i}$. Now, when we say that the WWW is a large collection of all kinds of stories, what we mean in particular is that not only $n\gg 1$, but that for many $i$ we will also have $n_{A,D_i}\approx n_{A,B}$, so that $n_{A,\bar{B}}\gtrsim n\, n_{A,B} \gg n_{A,B}$.

According to the above considerations, we obtain that in most situations of overextension we have $p(C|AB)\gg p(C|A\, {\rm then}\, B)$, and of course also $p(C|AB)\gg p(C|B\, {\rm then}\, A)$, i.e., the overextension effects  predicted by the EE model are typically at least one order of magnitude greater than those predicted by the OE model.

\section{The `Pet-Fish problem' and the `Linda story' on the WWW}\label{Guppy}
In this section, we apply the analysis of Sect. \ref{WWW} to two examples, to illustrate our results and to more specifically compare the orders of magnitude of the overextension effects produced by EE and OE. 

The first example is the so-called `Pet-Fish problem' of concept theory, also known as the `Guppy effect'. As suggested by Osherson and Smith \cite{os1981}, people rate the typicality of exemplars like \emph{Guppy} with respect to the conjunction \emph{Pet-Fish} as higher than their typicality with respect to both {\it Pet} and {\it Fish}. Thus, typicality judgments exhibit an unexpected behavior from the point of view of classical (fuzzy set) logic and probability theory.

To observe the equivalent of this behavior on the WWW, let us now make the three terms $A$, $B$ and $C$ of Sect. \ref{WWW} correspond to the terms \emph{Pet}, \emph{Fish}, and \emph{Guppy}, which we will denote by the letters $P$, $F$ and $G$, respectively. To estimate the numbers of pages containing these terms, and the specific  combination $PF$ = \emph{Pet-Fish}, we can use one of the available search engines, opting here for Google. 

It is important to keep in mind that counts are purely indicative and are just meant here to illustrate the results of the previous section, as is clear that the hits provided by Google are not always a precise measure of the actual number of pages existing on the WWW, for a given crtierion of search. 
In particular, depending on the nature of the query, Google will go more or less deep in its search for matches, and this can easily produce some logical inconsistencies. For instance, Google can return a number $n_A$ that is smaller than $n_{A,\bar{B}}$, which of course would be meaningless. An easy way to cope with this problem is to force Google to always search with the same deepness. For this, instead of looking for the number of pages containing the word $A$, we have used the trick of looking for the number of pages containing $A$, but excluding a given number of other words that we know in advance have no evident meaning relation with $A$, and can only produce very few matches when searched 
individually.\footnote{More precisely, in our searches for words and combination of words we have always excluded the following four very unusual Italian words: `barbablu', `miseriaccia', `acciderpoli' and `tristobello'.} 
In this way, we do not sensibly alter the counts for $A$, but since the search becomes more involved, Google will have to dig deeper and more consistent counts can be obtained. Proceeding in this way, on 
November 19, 
2016, a Google search has provided us with the following numbers: 
\begin{eqnarray}
&n_P=7.14\times 10^8,\, n_F=8.06\times 10^8,\, n_G=4.67\times 10^6,\nonumber\\
&n_{PF}=6.17\times 10^5,\, n_{P,F}=8.88 \times 10^7,\, n_{P,G}=6.12\times 10^5,\nonumber\\ 
&n_{F,G}=7.86\times 10^5,\, n_{PF,G}=1.37\times 10^5,\, n_{P,F,G}=6.70 \times 10^5,\nonumber\\ 
&n_{\bar{P},F}=7.57 \times 10^8,\, n_{P,\bar{F}}=6.52 \times 10^8.
\end{eqnarray}

According to (\ref{fallacy1}), we find for the EE model $p(G|PF)= 2.22 \times 10^{-1}$, $p(G|P)= 8.57 \times 10^{-4}$ and $p(G|F)= 9.75 \times 10^{-4}$, thus, a rather strong double overextension: 
\begin{equation}
{p(G|PF)\over p(G|P)}=2.59\times 10^2,\quad {p(G|PF)\over p(G|F)}=2.28\times 10^2.
\end{equation}
On the other hand, according to (\ref{conditional2}), we have $p(G|P\, {\rm then}\, F)= 9.38 \times 10^{-4}$ and $p(G|F\, {\rm then}\, P)= 8.31 \times 10^{-4}$, so we obtain the following 
very 
weak (single) overextension, if 
$P$ 
is considered first in the sequence: 
\begin{equation}
{p(G|P\, {\rm then}\, F)\over p(G|P)}=1.09.
\end{equation}
Comparing it with the overextension based on EE, we find:
\begin{equation}
{p(G|PF)\over p(G|P\, {\rm then}\, F)}=2.37 \times 10^{2},
\label{ratioEEversusOE}
\end{equation}
which is a difference of more than two orders of magnitude. 

It is instructive to also consider condition (\ref{condition}). We see that, in accordance with the heuristic reasoning we have presented in Sec.~\ref{comparing}, the first factor ${M(G; PF)\over M(G; P,F)}=2.94\times 10^{1}$ is much greater than $1$, and the same is true for the second factor $(1+{n_{P,\bar{F}}\over n_{P,F}})=8.34$, and if we calculate the ratio (\ref{ratioEEversusOE}) using expression (\ref{condition}), we find the very close value of $2.46\times 10^{2}$, showing that Google's counts were consistent when the `minus operator' was used to exclude words. 

The second example we want to consider is a traditional cognitive situation leading to the conjunction fallacy, known as the `Linda story' \cite{bb2012,tk1983}, where participants are presented with a questionnaire that includes the following narrative about a liberal woman named ``Linda'': \emph{``Linda is 31 years old, single, outspoken and very bright. She majored in philosophy. As a student, she was deeply concerned with issues of discrimination and social justice, and also participated in anti-nuclear demonstrations.''} 

They are then asked to check which of the two following alternatives is more likely: (i) Linda is a bank teller; (ii) Linda is a bank teller and is active in the feminist movement. Typically, the situation produces a conjunction fallacy, as respondents indicate on average that option (ii) is more likely than option (i). 

To adapt the `Linda story' to our WWW analysis (there are of course many ways to do that), we have considered the three terms \emph{Bank}, \emph{Feminism}, and \emph{Justice}, which we will denote by the letters $B$, $F$ and $J$, respectively. Again, a Google search performed on 
November 19, 
2016, has provided us with the following counts 
(always using the previously mentioned technique that consists of excluding certain terms, to obtain more consistent counts):
\begin{eqnarray}
&n_B=2.08\times 10^9,\, n_F=4.55\times 10^7,\, n_J=6.93\times 10^8, \nonumber\\
&n_{BF}=2.02\times 10^3,\, n_{B,F}=1.88 \times 10^6,\, n_{B,J}=1.86\times 10^8, \nonumber\\ 
&n_{F,J}=1.15\times 10^7,\, n_{BF,J}=8.32\times 10^2,\, n_{B,F,J}=1.31 \times 10^6,\nonumber\\ 
&n_{\bar{B},F}=4.67 \times 10^7,\, n_{B,\bar{F}}=2.08 \times 10^9.
\end{eqnarray}

We thus find for the EE model: $p(J|BF)= 4.12 \times 10^{-1}$, $p(J|B)= 8.94\times 10^{-2}$ and $p(J|F)= 2.53 \times 10^{-1}$, i.e., the double overextension: 
\begin{equation}
{p(J|BF)\over p(J|B)}=4.61,\quad {p(J|BF)\over p(J|F)}=1.63.
\end{equation}
On the other hand, we have $p(J|B\, {\rm then}\, F)= 6.30 \times 10^{-4}$ and $p(J|F\, {\rm then}\, B)= 2.88 \times 10^{-2}$, thus no overextension effects (conjunction fallacies) can be modeled in this case by means of OE. 

Let us however check once more the validity of the heuristic reasoning we have presented in Sec.~\ref{comparing}. The value of the first factor in (\ref{condition-1}) is now ${M(J;BF)\over M(J;B,F)}=5.91\times 10^{-1}$, i.e., of the order of $1$, in accordance with the fact that we are here in the presence of a much weaker 
form of 
fallacy, mostly associated with the joint presence of the two terms $B$ and $F$ 
in a same webpage, 
rather than with their presence in the exact $BF$-combination. Regarding the second factor in (\ref{condition-1}), its value is: $(1+{n_{B,\bar{F}}\over n_{B,F}})=1.11\times 10^{3}$, which is much greater than $1$, and if we calculate the ratio ${p(J|BF)\over p(J|B\, {\rm then}\, F)}$ using expression (\ref{condition}), we find a value of $6.55\times 10^{2}$, very close to the value of $6.54\times 10^{2}$ obtained using the alternative expression (\ref{condition-1}), showing again that Google's counts were consistent when the `minus operator' was used to exclude words.

\section{Conclusion}\label{Conclusion} 

In the present paper, we have emphasized that there are two major `quantum-based explanations' for the conjunction fallacy, as observed in psychological measurements. The first one is based on the hypothesis that respondents estimate conjunctions as a two-step reasoning that produces OE, responsible for the quantum interference whence the fallacy (and possibly other non-classical effects) arises \cite{pnas,Busemeyer2015a}. The second one is based on the hypothesis that respondents estimate a conjunction mostly by means of a single cognitive process (a single measurement), i.e., considering it a new emergent concept, with a meaning not reducible to that of its components \cite{asdbs2016,IQSA2,asdbs2016}. 

The advantage of the EE explanation is that it also accounts for double conjunction fallacies situations, whereas the OE approach is structurally unable to do so. Also, the EE approach does not require the presence of OE, and a recent work has indicated that OE are very weak in typical experimental contexts exhibiting conjunction fallacies \cite{bketal2015}.

In accordance with the analysis in \cite{bketal2015}, we have provided theoretical support to the EE model, by analyzing how overextension effects can appear on the WWW, when the latter is viewed as a 'mind-like entity that can tell a story'. For the sake of completeness, let us mention that the way of extracting conceptual structure from the WWW we have described in this article was put forward by one of us as part of the elaboration of a recent interpretation of quantum mechanics, called the `conceptuality interpretation' \cite{Aerts2009,Aerts2010}. 

An important ingredient in our analysis was the notion of `meaning bond' (\ref{bond1}), initially defined in \cite{a2011}. We mentioned already that the meaning bond is to be considered as a first-order approximation and foundation of a meaning bond measure. Indeed, more specific notions of `meaning bond' would attribute different weights to the webpages, or, to put it differently, would consider a reduced set of webpages in the calculation of the relative frequencies. For example, webpages where the references to the concepts we measure are negligible or misleading with respect to the content of the document and the nature of the `meaning relation' under consideration, should be eliminated from the calculation, whereas the webpages where these references are important and fundamental should be maintained.

This is probably what human subjects do, when they are asked to evaluate the likelihood of certain `meaning relations', like for instance the typicality of an exemplar with respect to some more abstract concepts and their combinations.  When they do so, they probably only consider stories that are felt to be relevant for the context in question. In other words, it is reasonable to assume that human minds have the tendency to renormalize their sets of stories in a way that only those that are perceived to be really relevant for the cognitive situation at hand are taken into account, and this may explain why the probabilities we have estimated by calculating relative frequencies of webpages appear to be rather small if compared to the estimations of probabilities usually operated by human subjects, which as is well known can easily incur in overestimations.

We conclude by observing that although proposals for more refined notions of meaning bound will rely on semantic technologies yet to be developed, the first-order `meaning bond' notion we have proposed might provide a starting point from which theories of subjective reasoning, usually relying on heuristics and ad-hoc assumptions, could not only be derived and understood from first principles, but also tested and evaluated using automatic processes.

\end{document}